\documentclass{article} 
\usepackage{iclr2022_conference,times}

\usepackage{amsmath,amsfonts,bm}

\def\eqref#1{equation~\ref{#1}}

\def\1{\bm{1}}

\DeclareMathAlphabet{\mathsfit}{\encodingdefault}{\sfdefault}{m}{sl}
\SetMathAlphabet{\mathsfit}{bold}{\encodingdefault}{\sfdefault}{bx}{n}

\usepackage{hyperref}
\usepackage{url}

\usepackage{booktabs}
\usepackage{subfigure}
\usepackage{graphicx}

\usepackage[noend]{algpseudocode}
\usepackage{algorithmicx,algorithm}
\usepackage{bm}

\title{Response-based Distillation for Incremental Object Detection}

\author{Tao Feng \\
Alibaba Group\\
\texttt{shisi.ft@alibaba-inc.com} \\
\And
Mang Wang \thanks{corresponding author}\\
Alibaba Group \\
\texttt{wangmang.wm@alibaba-inc.com} \\
}

\iclrfinalcopy 
\begin{document}

\maketitle

\begin{abstract}
Traditional object detection are ill-equipped for incremental learning. However, fine-tuning directly on a well-trained detection model with only new data will leads to catastrophic forgetting. Knowledge distillation is a straightforward way to mitigate catastrophic forgetting. In Incremental Object Detection (IOD), previous work mainly focuses on feature-level knowledge distillation, but the different response of detector has not been fully explored yet. In this paper, we propose a fully response-based incremental distillation method focusing on learning response from detection bounding boxes and classification predictions. Firstly, our method transferring category knowledge while equipping student model with the ability to retain localization knowledge during incremental learning. In addition, we further evaluate the qualities of all locations and provides valuable response by adaptive pseudo-label selection (APS) strategies. Finally, we elucidate that knowledge from different responses should be assigned with different importance during incremental distillation. Extensive experiments conducted on MS COCO demonstrate significant advantages of our method, which substantially narrow the performance gap towards full training.
\end{abstract}

\section{Introduction}

In the natural world, the visual system of creatures could constantly acquire, integrate and optimize knowledge. Learning mode is inherently incremental for them. In contrast, currently, the classic training paradigm of the object detection model \citep{DBLP:conf/iccv/TianSCH19,DBLP:conf/cvpr/LiW0LT021} does not have such capability. Supervised object detection paradigm relies on accessing pre-defined labeled data. This learning paradigm implicit assumes data distribution is fixed or stationary, while data from real world is represented by continuous and dynamic data flow, whose distribution is non-stationary. When the model continuously obtains knowledge from non-stationary data distribution, new knowledge would interfere with the old one, triggering catastrophic forgetting \citep{goodfellow2015empirical, 1989Catastrophic}.

A straightforward way in incremental object detection is based on knowledge distillation \citep{DBLP:journals/corr/HintonVD15}. \citet{DBLP:journals/cviu/PengZMLL21} stressed that the Tower layers could reduce catastrophic forgetting significantly. They implemented incremental learning on an anchor-free detector and selectively performed distillation on non-regression outputs. In knowledge distillation for object detection where incremental learning was not introduced, previous work extracted knowledge from the combined distillation of different components. For example, \citet{DBLP:conf/nips/ChenCYHC17} and \citet{DBLP:journals/corr/abs-2006-13108} distilled all components of the detector. Nevertheless, the nature of these methods are designed using feature-based knowledge distillation \citep{DBLP:conf/cvpr/Chen0ZJ21}, fully response-based method \citep{DBLP:journals/ijcv/GouYMT21} has not been explored in incremental object detection \citep{DBLP:journals/tnn/LiuKCXYZ21} yet. Besides, since different components in the detection make different contributions to incremental distillation, an elaborate design for different responses is essential \citep{DBLP:conf/iccv/LinGGHD17}.

This paper focused on a practical and challenging problem concerning incremental object detection: \emph{how to learn response from detecting bounding boxes and classification predictions}. Responses in object detection contain logits together with the offset of bounding box \citep{DBLP:journals/ijcv/GouYMT21}. Firstly, since the number of ground truth on each new image is uncertain, one of the foremost considerations is that validate the object of all samples, determining which object is positive or negative and which ground truth each object should regress towards. A troublesome issue is that the output of the regression branch may be substantially different from that of the ground truth. Furthermore, the localization knowledge of each edge in the detection bounding boxes is also response that should be taken seriously. To sum up, we use the response on the location where teacher detector generates high-quality predictions as the ground truth to guide the student detector following the behavior of teacher on the old object. In this case, it is of great significance to use the old detector to provide valuable incremental information from detection bounding boxes and classification predictions.

To tackle the above problems, this paper rethinks response-based knowledge distillation method, finding that distillation at proper locations is crucial in facilitating incremental object detection. We believe that student detector can acquire high-quality knowledge from the teacher detector’s high-quality predictions. Driven by this inspiration, we proposed an incremental distillation scheme that learns specific responses from the classification head and regression head respectively. Unlike previous work, we introduce incremental localization distillation \citep{zheng2021localization} in regression response to equip student detector with the ability to learn location ambiguity \citep{DBLP:conf/nips/0041WW00LT020} during incremental learning. Besides, we propose adaptive pseudo-label selection (APS) strategies to automatically select distillation nodes based on statistical characteristics from different responses, which evaluates the qualities of all locations and provides valuable response. We alleviate catastrophic forgetting greatly and significantly narrow the gap with full training by distilling the response alone. Extensive experiments on the MS COCO dataset support our analysis and conclusion. 

The main contributions of this work can be summarized,

\begin{enumerate} 
\item To the best of our knowledge, this paper is first work to explore the fully response-based distillation method in incremental object detection.
\item We propose a novel distillation scheme elaborate designed for incremental detection focusing on detection bounding boxes and classification predictions.
\item We propose adaptive pseudo-label selection strategies to automatically select distillation nodes based on statistical characteristics from the different responses.
\end{enumerate}

\section{Related work}

\textbf{Incremental Learning.} Catastrophic forgetting is the core challenge for incremental learning. Incremental learning based on parameter constraints is a candidate solution for such problem, which protects the old knowledge by introducing an additional parameter-related regularization term to modify the gradient. EWC \citep{DBLP:journals/corr/KirkpatrickPRVD16} and MAS \citep{DBLP:conf/eccv/AljundiBERT18} are two typical representatives of such method. Another solution is incremental learning based on knowledge distillation, as well as the topic of the study. This kind of method mainly projects old knowledge by transferring knowledge in old tasks to new tasks through knowledge distillation. LwF \citep{DBLP:journals/pami/LiH18a} is the first algorithm that introduces the concept of knowledge distillation into incremental learning, in the purpose of making predictions of the new model on new tasks similar to that of the old model and thereby protecting the old knowledge in the form of knowledge transfer. However, it would cause knowledge confusion when the correlation between new and old tasks is low. iCaRL \citep{DBLP:conf/cvpr/RebuffiKSL17} algorithm uses knowledge distillation to avoid excessive deterioration of knowledge in the network, while BiC \citep{DBLP:conf/cvpr/WuCWYLGF19} algorithm added a bias correction layer after the FC layer to offset the category bias of new data when using the distillation loss.

\textbf{Incremental Object Detection.} Compared with incremental classification, achievements on incremental object detection is much less. Meanwhile, the high complexity of the detection task also adds the difficulty of incremental object detection. \citet{DBLP:conf/iccv/ShmelkovSA17} proposed to apply LwF to Fast RCNN detector \citep{girshick2015fast}, which is the first work on incremental object detection. Thereafter, some researchers move this area forward. \citet{DBLP:journals/cviu/PengZMLL21} proposed SID approach for incremental object detection on anchor-free detector and conducted experiments on FCOS \citep{DBLP:conf/iccv/TianSCH19} and CenterNet \citep{zhou2019objects}. \citet{li2021classincremental} studied object detection based on class-incremental learning on Faster RCNN detector with emphasis given to few-shot scenarios, which is also the focus of ONCE algorithm \citep{DBLP:conf/cvpr/Perez-RuaZHX20}. \citet{DBLP:conf/edge/LiTGZZH19} designed an incremental object detection system with RetinaNet detector \citep{DBLP:journals/pami/LinGGHD20} under the scenario of edge device. the latest work, \citet{DBLP:conf/cvpr/Joseph0KB21} introduced the concept of incremental learning when defining the problems of Open World Object Detection (OWOD).

\textbf{Knowledge Distillation for Object Detection.} Knowledge distillation \citep{DBLP:conf/kdd/BucilaCN06}is an effective way to transfer knowledge between models. Widely applied in image classification tasks in previous researches, knowledge distillation is now used in object detection tasks more and more frequently. \citet{DBLP:conf/nips/ChenCYHC17} implemented distillation in all components on Faster RCNN detector (including backbone, proposals in RPN, and head). To imitate the high-level feature response of the teacher model with the student model, \citet{DBLP:conf/cvpr/WangYZF19} proposed a distillation method based on fine-grained feature imitation. By synthesize category-conditioned objects through inverse mapping, \citet{DBLP:conf/wacv/ChawlaYMA21} proposed a data-free knowledge distillation technology applicable for object detection, but the method would trigger dream-image. \citet{DBLP:conf/cvpr/Guo00W0X021} believing that foreground and background both play an unique role in object detection, proposed an object detection distillation method that could decouple foreground and background. \citet{zheng2021localization} proposed a localization distillation method introducing knowledge distillation into the regression branch of the object detector, so as to enable the student network to solve the localization ambiguity in object detection as the teacher network. However, existing object detection distillation framework does not pay enough attention to the significant role of the head. In this study, we found the head has its particularly significant.

\begin{figure}
\centering
\includegraphics[width=1\textwidth]{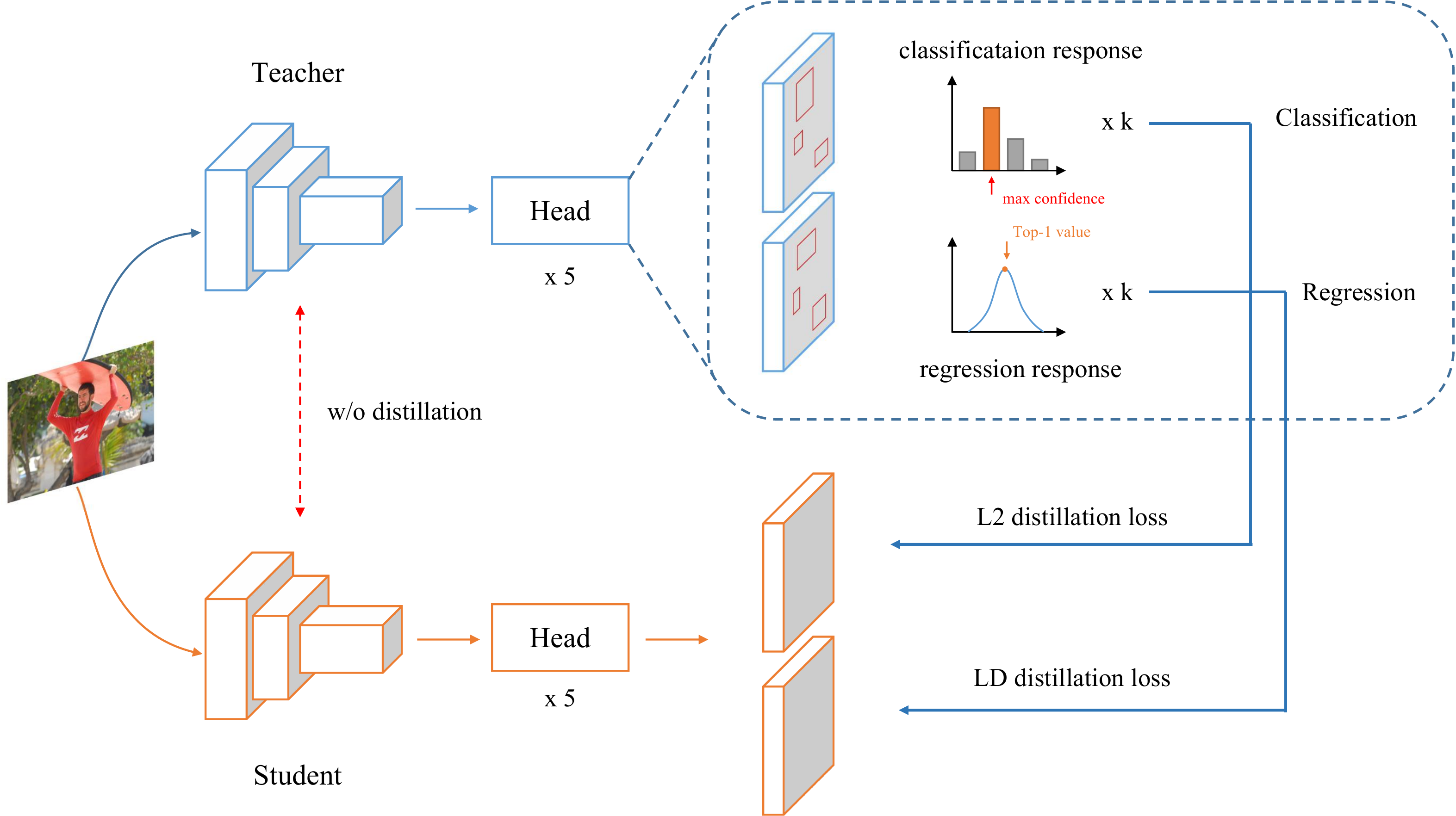}
\caption{Overview of response-based incremental distillation.}
\label{figure1}
\end{figure}

\section{Method}

\subsection{Overall Structure}

In general, a one-stage object detector \citep{DBLP:journals/corr/abs-1905-05055,DBLP:conf/nips/0041WW00LT020}is composed of three components: (\romannumeral1.) backbone for feature extraction; (\romannumeral2.) neck for fusion of multi-level features; (\romannumeral3.) head for classification and regression. The purpose of incremental distillation is to transfer old knowledge to the student detector, and this knowledge could be the features of the intermediate layer in the backbone or neck or the soft predictions in the head. Here, we incrementally learn a strong and efficient student object detector by the distillation of incremental knowledge from responses of the different heads. The overall incremental detection framework is shown in Figure \ref{figure1}. Firstly, knowledge distillation is applied to learn incremental response from the classification head and regression head of the teacher detector. Secondly, incremental localization distillation loss is also applied to enhance the localization information extraction ability of the student detector. Notably, the adaptive pseudo-label selection strategies are proposed to gain more meaningful incremental responses from the teacher detector, that is, selective calculation of the distillation loss from the pseudo label provided by the teacher detector. The overall learning target of the student detector is therefore defined as,

\begin{equation}
\mathcal{L}_{{total}}=\mathcal{L}_{model}+\lambda_{1} \mathcal{L}_{dist\_{cls}}(\mathcal{{C}_{S}}, \mathcal{{C}_{T}})+\lambda_{2} \mathcal{L}_{dist\_{bbox}}(\mathcal{{R}_{S}}, \mathcal{{R}_{T}})
\end{equation}

where $\lambda$ is the parameters that balances the weights of different loss terms. The loss term $\mathcal{L}_{model}$ is standard loss function used in GFLV1 \citep{DBLP:conf/nips/0041WW00LT020} to train object detector for the new object class. The second loss term $\mathcal{L}_{dist\_{cls}}$ is the L2 incremental distillation loss for classification branch. The third loss term $\mathcal{L}_{dist\_{bbox}}$ is the incremental localization distillation loss for regression branch. In the above, we set $\lambda_{1}=\lambda_{2}=1$.

\subsection{Distillation at Classification-based Response}

The soft predictions from the classification head contains the knowledge of various categories discovered by the teacher model. Through the learning of soft prediction, the student model can inherit hidden knowledge, which is intuitive for classification tasks.  Let $\mathcal{T}$ be the teacher model, we use SoftMax function to transform logits $\mathcal{{Z}_{T}}$ in final score output, responding probability distribution $\mathcal{{P}_{T}}$ is defined as,

\begin{equation}
\mathcal{{P}_{T}}=\operatorname{SoftMax}\left(\frac{\mathcal{{Z}_{T}}}{t}\right)
\end{equation}

Similarly, we define $\mathcal{{P}_{S}}$ for the student model $\mathcal{S}$,

\begin{equation}
\mathcal{{P}_{S}}=\operatorname{SoftMax}\left(\frac{\mathcal{{Z}_{S}}}{t}\right)
\end{equation}

where $t$ is the temperature to soften the probability distribution for $\mathcal{{P}_{T}}$ and $\mathcal{{P}_{S}}$.

Previous works usually directly use all the prediction responses in the classification head and treat each position equally. If there is any inappropriate balance, the response generated by the background category may overwhelm the response generated by the foreground category, thereby interfering with the retention of old knowledge. To tackle this problem, the L2 incremental distillation loss for the classification-based response is as follows,

\begin{equation}
\mathcal{L}_{{dist\_{cls}}}\left(\mathcal{{C}_{S}}, \mathcal{{C}_{T}}\right)=\sum_{i=1}^{m}\left(\mathcal{{P}_{S}}^{i}-\mathcal{{P}_{T}}^{i}\right)^{2}
\label{eq4}
\end{equation}

where $\mathcal{{P}_{T}}^{i}$  is the category response of the frozen teacher detector from $m$ selected pseudo object classes using the new data, and $\mathcal{{P}_{S}}^{i}$ is the corresponding category response of the student detector. By distilling the selected response, the student detector inherits the knowledge of the positive object category to a greater extent.

\subsection{Distillation at Regression-based Response}

The bounding box response from the regression branch is also quite important for incremental detection. Contrary to the discrete class information, there is a possibility that the output of the regression branch may provide a regression direction that contradicts the ground truth. That’s because, even if the image does not contain any objects of the old category, the regression branch will still predict the bounding box, although the confidence is relatively low. That poses a challenge for learning the knowledge of the old model to correctly predict the bounding box of the old object. On the other hand, in previous works, only the bounding box of a relatively high-confidence object was learned as the knowledge of the teacher detector, ignoring the localization information.

Benefit from the general distribution of bounding box $\mathcal{B}$ from GFLV1 detector, each edge of $\mathcal{B}$ can be represented by probability distribution through SoftMax function \citep{zheng2021localization}. Thus, the probability matrix of bounding box $\mathcal{B}$ can be defined as, 

\begin{equation}
\mathcal{B}=\left[p_{t}, p_{b}, p_{l}, p_{r}\right] \varepsilon \mathbb{R}^{n \times 4}
\end{equation}

Therefore, we can extract the incremental localization knowledge of bounding box $\mathcal{B}$ from the teacher detector $\mathcal{T}$ and transfer it to the student detector $\mathcal{S}$ by using KL-Divergence loss, 

\begin{equation}
\mathcal{L}_{L D}^{e}=\mathcal{L}_{K L}\left(\mathcal{{P}_{S}}^{j}, \mathcal{{P}_{T}}^{j}\right)
\end{equation}

Finally, incremental localization distillation loss for the regression-based response is defined as,

\begin{equation}
\mathcal{L}_{{dist\_{bbox}}}\left(\mathcal{{R}_{S}}, \mathcal{{R}_{T}}\right)=\sum_{j=1}^{J} \sum_{e \in \mathcal{B}} \mathcal{L}_{L D}^{e}
\label{eq7}
\end{equation}

where $\mathcal{{P}_{T}}^{j}$  is the regression response of the frozen teacher detector from $J$ pseudo bounding box using the new object, and $\mathcal{{P}_{S}}^{j}$ is the regression response of the student detector for the old bounding box. Compared to only use classification information in previous works, incremental localization distillation can provide extra of localization.

\subsection{Adaptive Pseudo-label Selection}

When an incremental object detector is trained, the gap of knowledge between the teacher detector and the student detector is obvious. For a new sample, it’s preferable for the teacher detector to provide the high-quality knowledge, as the student detector will benefit from positive response. To this end, a basic problem related to incremental object detection has been thoroughly studied: \emph{how to select distillation nodes as positive response.} Traditional selection strategies depend on sensitive hyper-parameters such as setting confidence or selecting Top-K . These empirical practices in which rules are fixed have such consequences that too small thresholds lead to the ignoring of some objects while too large ones result in the introduction of negative response.

To solve this problem, the adaptive pseudo-label selection (APS) strategy is proposed. Algorithm \ref{algorithm1} describes how the proposed strategy works for an input image. We obtain positive response from the category and bounding box as distillation nodes respectively.

\textbf{Classification head.} The statistical characteristics of the category information are utilized to determine the response of classification, as described in L-3 to L-11. We first calculate the classification confidence of each position. After that, we calculate the mean $\mu_{C}$ and standard deviation $\sigma_{C}$ in L-5 and L-6. With these statistical, the threshold $\tau_{C}$ is obtained in L-7. Finally, we select these candidates whose confidence are greater than the threshold $\tau_{C}$ in L-8 to L-11.

\textbf{Regression head.} The statistical characteristics of the distribution information are utilized to determine the response of regression, as described in L-13 to L-22. For the GFLV1 \citep{DBLP:conf/nips/0041WW00LT020} detector, the author points out that a certain and unambiguous bounding box, the distribution is usually sharp. Therefore, the Top-1 value is usually very large if the distribution is sharp. Based on these statistical characteristics, the Top-1 value is used to measure the confidence of the bounding box. Specifically, we first calculate the Top-1 value of each distribution. After that, we calculate the mean $\mu_{B}$ and the standard deviation $\sigma_{B}$ of all Top-1 value in L-15 and L-16. Then, the threshold  $\tau_{B}$ is obtained in L-17. Finally, we select these candidates whose confidence are greater than the threshold  $\tau_{B}$ in L-18 to L-20. The \emph{nms} operator returns a sampled set that is filtered using NMS.  

The proposed APS strategy has the following advantages: 1. guaranteeing fair selection of pseudo labels of different objects. 2. using statistical characteristics of different branches to adaptively select pseudo labels to provide the incremental response.

\begin{algorithm}[t]
\caption{Adaptive Pseudo-label Selection (APS)}
\label{algorithm1}
\hspace*{0.02in} {\bf Input:}
Unlabeled image \bm{$I$}, image-level labels \bm{$c$}, \bm{$b$}, teacher detector \bm{$\theta^{\prime}$} \\
\hspace*{0.02in} {\bf Output:}
Sampled pseudo-label sets \bm{$C^{\prime}$}, \bm{$B^{\prime}$}
\begin{algorithmic}[1]
\State Inference \bm{$I$} with \bm{$\theta^{\prime}$} yields the classification score \bm{$C$} and predicted distribution \bm{$B$} \\

\Statex \emph{// Classification branch}
\For{$k=1$ to $C$} 
    \State $G_{C}$ $\longleftarrow$ ${confidence(C_{k})}$
\EndFor
\State Compute $\mu_{C}=mean(G_{C})$
\State Compute $\sigma_{C}=std(G_{C})$
\State Compute threshold $\tau_{C}=\mu_{C} + \alpha \sigma_{C}$
\For{each candidate $c$ in $C$}
    \If{$G_{C_k} \geq \tau_{C}$}
    \State Add candidate $c$ to \bm{$C^{\prime}$}
    \EndIf
\EndFor
\State \Return \bm{$C^{\prime}$} \\

\Statex \emph{// Regression branch}
\For{$k=1$ to $B$} 
    \State $G_{B}$ $\longleftarrow$ ${\operatorname{Top-1}(B_{k})}$
\EndFor
\State Compute $\mu_{B}=mean(G_{B})$
\State Compute $\sigma_{B}=std(G_{B})$
\State Compute threshold $\tau_{B}=\mu_{B} + \alpha \sigma_{B}$
\For{each candidate $b$ in $B$}
    \If{$G_{B_k} \geq \tau_{B}$}
    \State Add candidate $b$ to \bm{$B^{\prime}$}
    \EndIf
\EndFor
\State \bm{$B^{\prime}$} $\longleftarrow$ $nms(B^{\prime},G_{B})$

\State \Return \bm{$B^{\prime}$}
\end{algorithmic}
\end{algorithm}

\section{Experiments and Discussion}

In this section, we perform experiments on several incremental scenarios on the MS COCO dataset using baseline detector GFLV1 to validate our method. Then, we perform ablation studies to prove the effectiveness of each component of our method. Finally, we discuss the application scenario in our method.

\textbf{Implementation Details.} We build our method on top of the GFLV1 \citep{DBLP:conf/nips/0041WW00LT020} detector using their official implementations. The teacher and student detectors defined in our experiments are standard GFLV1 architectures. For the GFLV1 detector, ResNet-50 is used as its backbone, FPN \citep{DBLP:conf/cvpr/LinDGHHB17} is used as its neck. We trained our detector to follow the same parameters described in their paper. All the experiments are performed on 8 NVIDIA Tesla V100 GPU, with batch size of 8.

\textbf{Datasets and Evaluation Metric.} MS COCO 2017 \citep{DBLP:journals/corr/ChenFLVGDZ15} is a challenging benchmark in object detection which contains 80 object classes. For experiments on the COCO dataset, we use train and validation set for training and test set for testing. The standard COCO protocols are used as an evaluation metric, i.e. $AP$, $AP_{50}$, $AP_{75}$, $AP_{S}$, $AP_{M}$ and $AP_{L}$. 

\textbf{Experiment Setup for MS COCO.} The detector is trained by 12 epochs (1x mode) for each incremental step for the MS COCO dataset. The setting is consistent for all the detectors in the different scenarios. We set up experiments in the following scenarios:
\begin{itemize}
\item \textbf{ 40 + 40:} we train a base detector with the first 40 classes and then the last 40 classes are learned incrementally as new object classes.
\item \textbf{75 + 5:} we train a base detector with the first 75 classes and then the last 5 classes are learned incrementally as new object classes.
\item \textbf{Last 40 + First 40:} we specially train a base detector with the last 40 classes and then the first 40 classes are learned incrementally as new object classes.
\end{itemize}

\begin{table}[]
\centering
\caption{Incremental results based on GFLV1 detector on COCO benchmark under first 40 classes + last 40 classes. ("$\Delta$" represents an improvement over Catastrophic Forgetting. "$\nabla$" represents the gap with Upper Bound.)}
\label{table1}
\begin{tabular}{@{}l|cccccc@{}}
\toprule
Method & $AP$ & $AP_{50}$ & $AP_{75}$  & $AP_{S}$  & $AP_{M}$  & $AP_{L}$  \\ \midrule
Upper Bound & 40.2 & 58.3  & 43.6  & 23.2  & 44.1   & 52.2  \\
Catastrophic Forgetting & 17.8 & 25.9  & 19.3 & 8.3  & 19.2 & 24.6\\ \midrule
LwF \citep{DBLP:journals/pami/LiH18a} & 17.2($\Delta -0.6/\nabla 23.0$) & 25.4  & 18.6 & 7.9  & 18.4 & 24.3\\
RILOD \citep{DBLP:conf/edge/LiTGZZH19} & 29.9($\Delta 12.1/\nabla 10.3$) & 45.0  & 32.0 & 15.8  & 33.0 & 40.5\\
SID \citep{DBLP:journals/cviu/PengZMLL21} & 34.0($\Delta 16.2/\nabla 6.2$) & 51.4  & 36.3 & 18.4  & 38.4 & 44.9\\
Ours  & \textbf{36.9($\Delta 19.1/\nabla 3.3$)} & \textbf{54.5}  & \textbf{39.6}  & \textbf{21.3}  & \textbf{40.4}   & \textbf{47.5}\\ \bottomrule
\end{tabular}
\end{table}

\begin{table}[]
\centering
\caption{Incremental results based on GFLV1 detector on COCO benchmark under last 40 classes + first 40 classes. ("$\Delta$" represents an improvement over Catastrophic Forgetting. "$\nabla$" represents the gap with Upper Bound.)}
\label{table2}
\begin{tabular}{@{}l|cccccc@{}}
\toprule
Method & $AP$ & $AP_{50}$ & $AP_{75}$  & $AP_{S}$  & $AP_{M}$  & $AP_{L}$  \\ \midrule
Upper Bound & 40.2 & 58.3  & 43.6  & 23.2  & 44.1   & 52.2  \\
Catastrophic Forgetting & 22.6 & 32.7  & 24.2 & 15.1  & 25.0 & 27.6\\ \midrule
LwF \citep{DBLP:journals/pami/LiH18a} & 20.5($\Delta -2.1/\nabla 19.7$) & 29.9  & 22.1 & 13.0  & 22.5 & 25.3\\
RILOD \citep{DBLP:conf/edge/LiTGZZH19} & 34.1($\Delta 11.5/\nabla 6.1$) & 51.1  & 36.8 & 19.1  & 38.0 & 43.9\\
SID \citep{DBLP:journals/cviu/PengZMLL21} & 33.5 ($\Delta 10.9/\nabla 6.7$) & 50.9  & 36.3 & 19.0  & 37.7 & 43.0\\
Ours & \textbf{37.5($\Delta 14.9/\nabla 2.7$)} & \textbf{55.1}  & \textbf{40.4}  & \textbf{21.3}  & \textbf{41.1}   & \textbf{48.2}\\ \bottomrule
\end{tabular}
\end{table}

\subsection{Overall Performance}

\begin{figure*}
\centering
\subfigure[Upper Bound]{%
\includegraphics[width=1.7in]{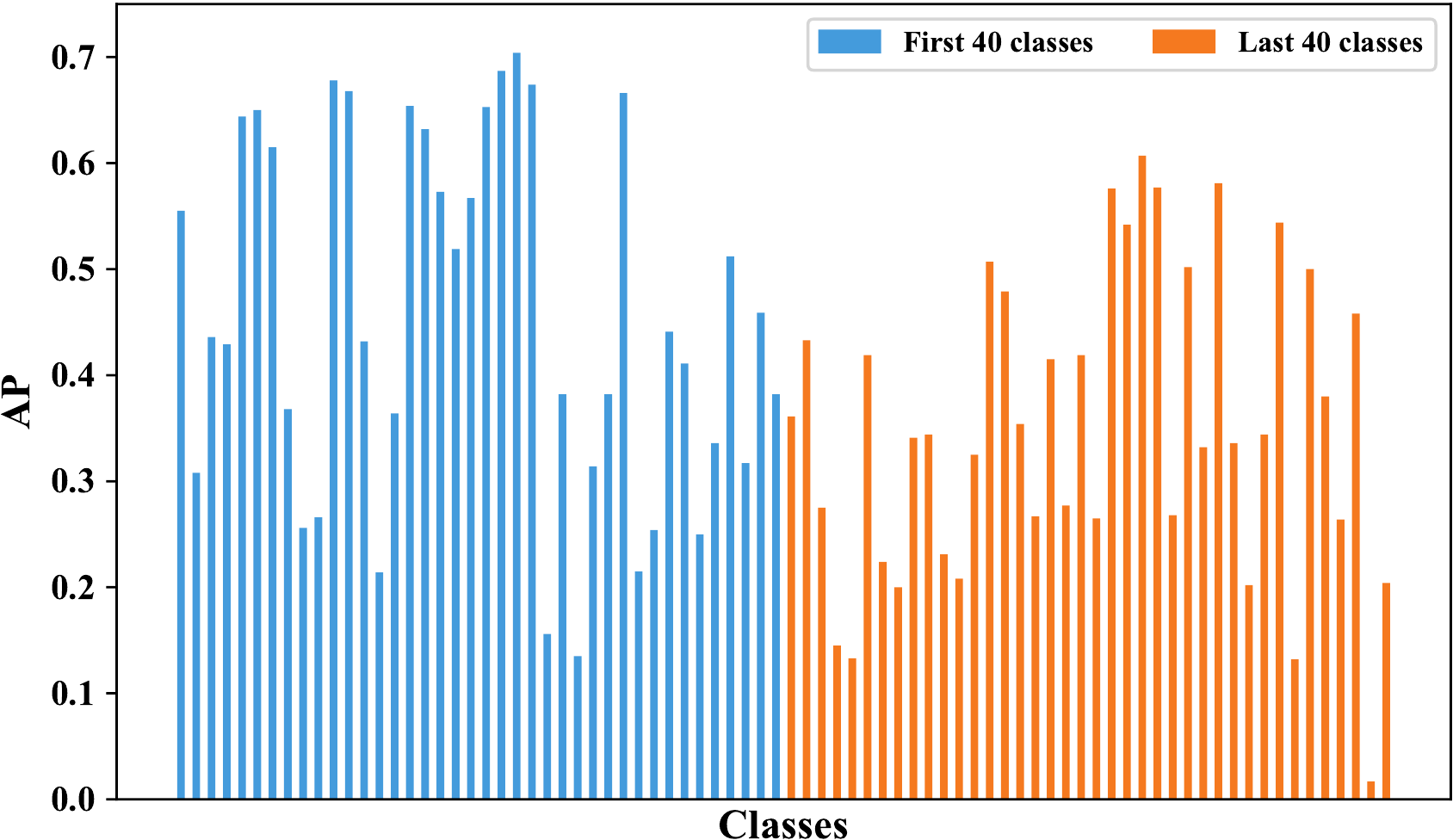}
\label{fig:subfigure21}}
\quad
\subfigure[Catastrophic Forgetting]{%
\includegraphics[width=1.7in]{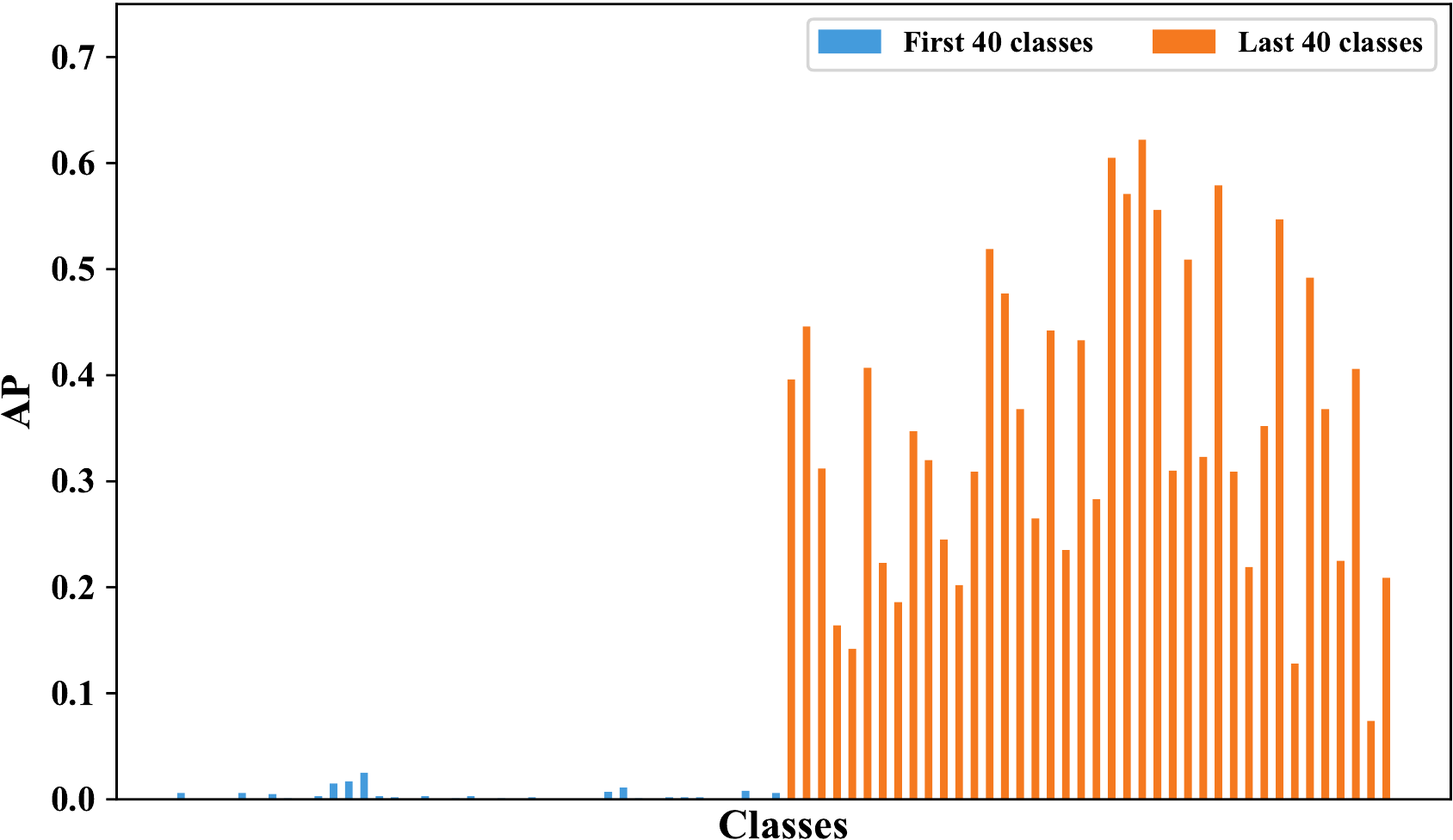}
\label{fig:subfigure22}}
\subfigure[Response-based Distillation]{%
\includegraphics[width=1.7in]{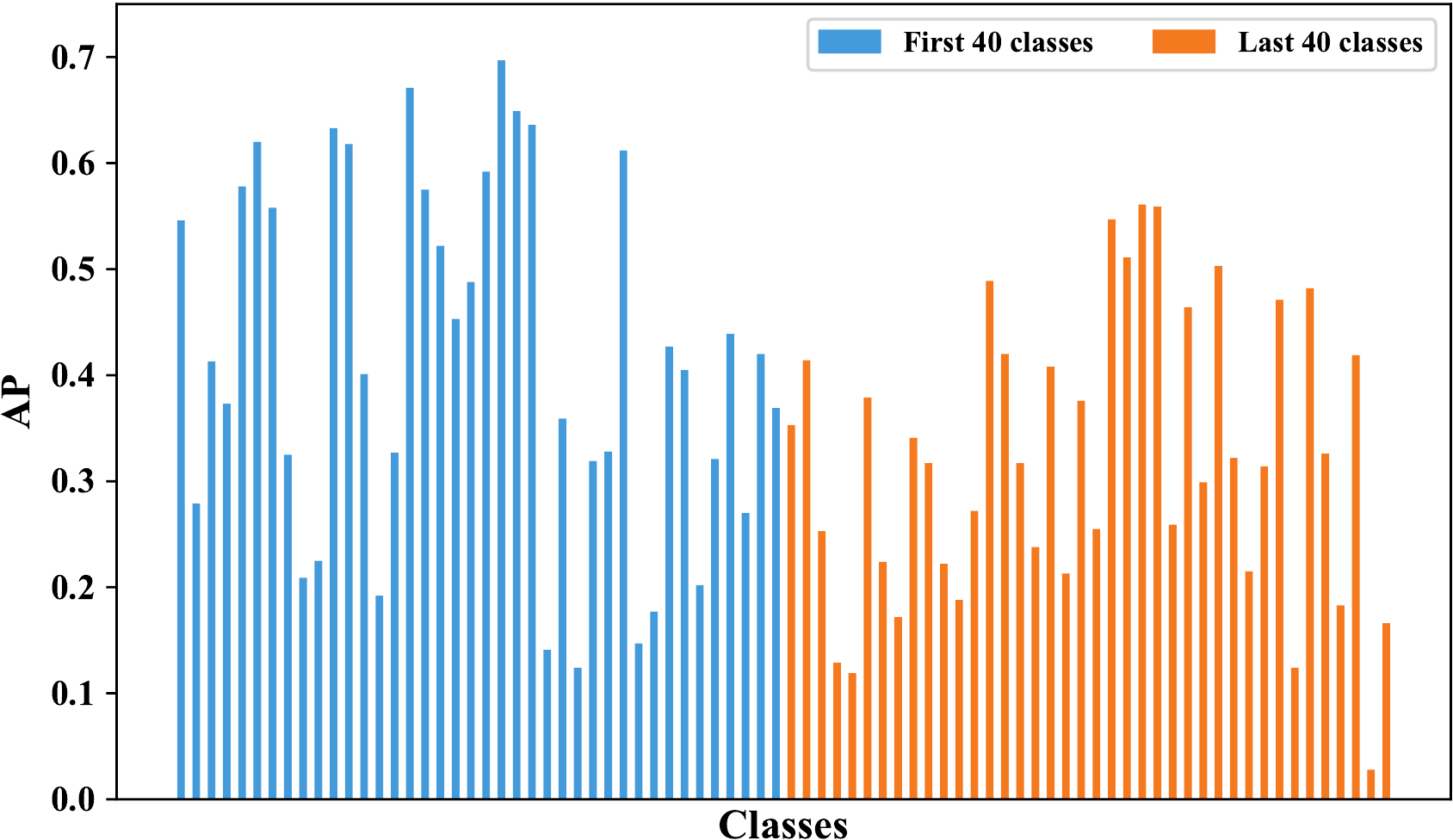}
\label{fig:subfigure23}}
\caption{AP of Per-class among different learning schemes. (a) Detector is trained with all data.(b) Student detector is finetuned with new classes.(c) Student detector is distilled via different response.}
\label{figure2}
\end{figure*}

\begin{figure}[htbp]
\centering
\begin{minipage}[t]{0.48\textwidth}
\centering
\includegraphics[width=4.9cm]{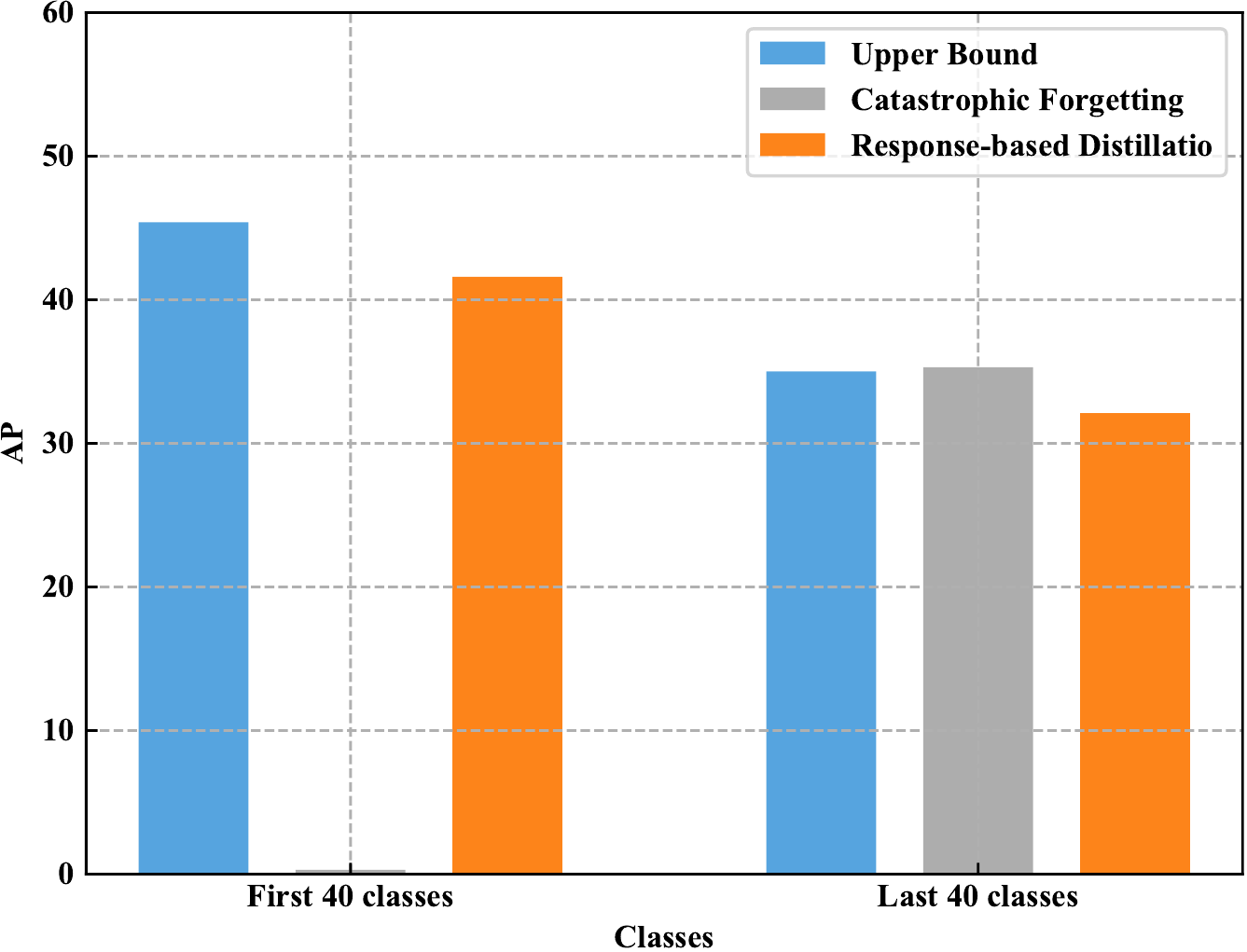}
\caption{First 40 classes vs. Last 40 classes.}
\label{figure3}
\end{minipage}
\begin{minipage}[t]{0.48\textwidth}
\centering
\includegraphics[width=5cm]{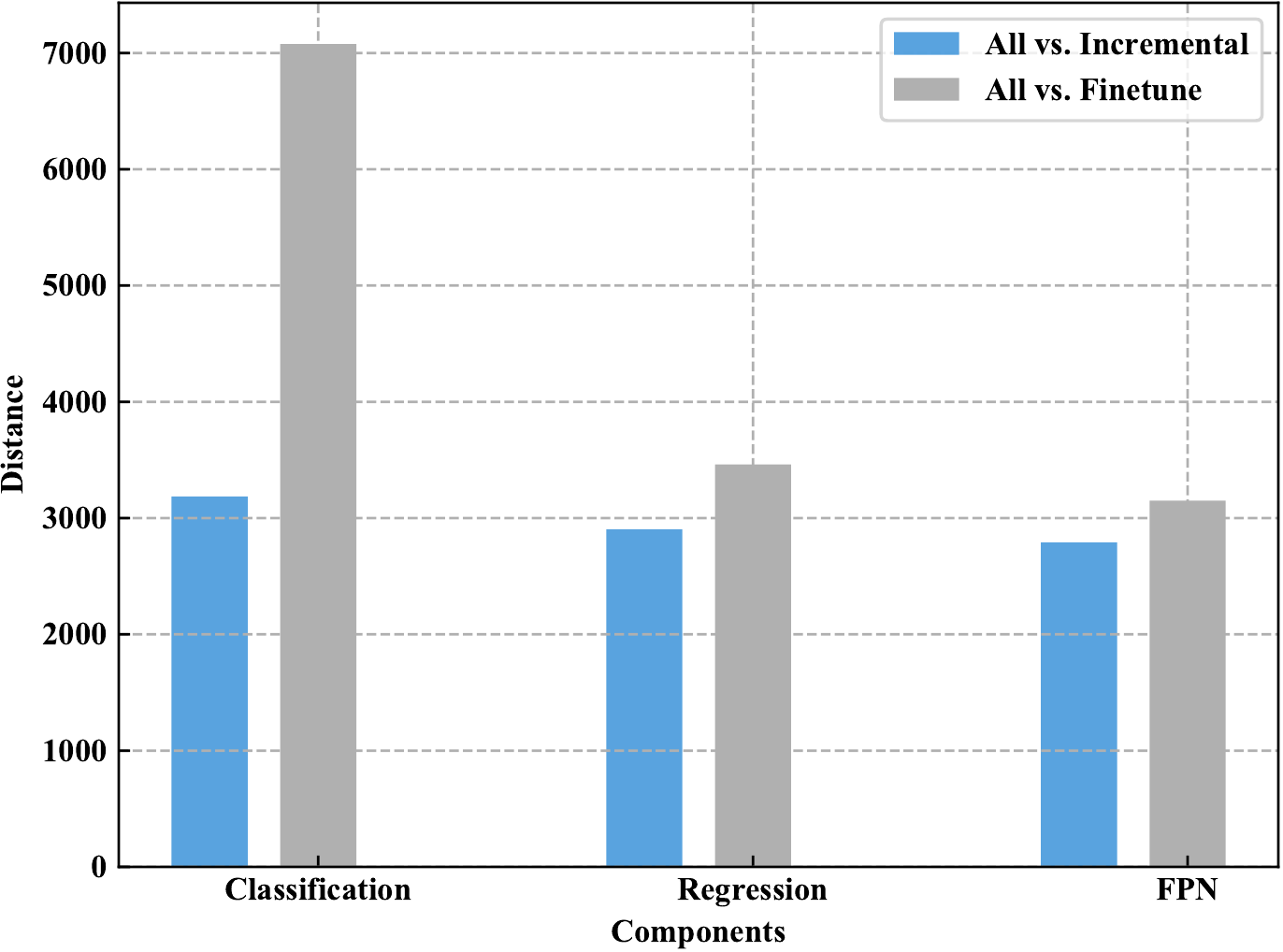}
\caption{L2 distance analysis.}
\label{figure4}
\end{minipage}
\end{figure}

We reported the incremental results under the first 40 classes + last 40 classes scenario in Table \ref{table1}. In this scenario, we observed that if the old detector and new data were directly used to conduct fine-tuning process, then the AP dropped to 17.8\% as compared to the 40.2\% in full data training. This is because the fine-tuning made the detector's memory of old object classes close to 0, resulting in catastrophic forgetting (ref to Figure \ref{fig:subfigure22}). Our method far outperformed fine-tuning across various IoUs evaluation criteria from 0.5 to 0.95. The experimental results show that when IoU is 0.5, 0.75 and 0.95, the AP improves by 19.1\%, 28.6\% and 20.3\%, respectively. This indicates that our method can well address catastrophic forgetting. Notably, even compared with the full data training where the entire dataset was used, our method only had a gap of 3.3\%. This indicates that the student detector maintained a good memory of the old objects while learning new objects. To put it more intuitively, we visualized the incremental results of all object classes, as shown in Figure \ref{figure2}. The blue column denotes the AP of the first 40 classes, while the orange column denotes the AP of the last 40 classes. As can be seen, our method has produced significant outcomes. In Figure \ref{figure3}, we further visualized the AP of all objects of the first 40 classes and the last 40 classes.

Considering the long-tail problem of the COCO dataset, we particularly configured an incremental experiment under the last 40 classes + first 40 classes scenario. In this scenario, the first 40 classes object contain more memories that should be retained, which means that more incremental responses can be obtained. As can be seen from Table \ref{table2}, the incremental performance of our method has been further improved, with the gap against full data training reduced to 2.7\% and the improvement on catastrophic forgetting increased to 14.9\%. This also validates our inference that the method we propose benefits from more incremental responses. 

In addition, we also compared our method with LwF \citep{DBLP:journals/pami/LiH18a}, RILOD \citep{DBLP:conf/edge/LiTGZZH19}, and SID \citep{DBLP:journals/cviu/PengZMLL21}. Both Table \ref{table1} and Table \ref{table2} show that although LwF works well in incremental classification, it is even lower AP than directly fine-tuning in detection tasks. To a fair comparison with RILOD and SID, we replicated them based on GFLV1 detector. For RILOD, we completely followed their method. For SID, we used the component with the greatest improvement proposed by the authors. Both tables show that the improvement of our method to catastrophic forgetting is outstanding.

\subsection{Ablation Study}

As shown in Table \ref{table3}, we validated the effectiveness of different components of the proposed method on the COCO benchmark to highlight our improvement in performance. “all cls + all reg” denotes that responses from both the classification branch and regression branch are treated equally in the incremental distillation, which is also our baseline performance. “all cls” denotes that only classification responses in the incremental distillation process are treated equally. “all reg” denotes that only regression responses in the incremental distillation process are treated equally. “cls + APS” denotes that the APS strategy is employed to conduct incremental distillation over classification responses, as shown in Equation \ref{eq4}. “cls + reg +APS” denotes that responses based on regression are also used, as shown in Equation \ref{eq7}. In Table \ref{table3}, separately distillation all responses from classification and regression, obtained 23.8\% and 13.0\% of AP. When only all responses from the regression branch are used, AP is even lower than the fine-tuning performance, which also supports our assumption stated in the introduction section. Comparatively, the direct incremental distillation of all responses from classification and regression branches obtains 31.5\% of AP. By utilizing APS to decouple classification responses, the student detector obtained higher results. Our decoupling proposal can improve the result from 31.5\% of AP to 33.2\%. The incremental distillation process further utilized the APS strategy to decouple regression responses, obtaining 36.9\% of AP on the COCO benchmark, a 5.4\% improvement compared with the baseline performance. All these results clearly point to the advantageous performance of our method.

\subsection{Discussion}

In this section, we present further insights into response-based incremental distillation. We reveal the contribution of different components for distillation detection and discuss the impact of incremental response in the head.

\textbf{Distance between different components.} We calculate the feature distance between different components to illustrate why response-based distillation can attain higher performance compared to other components. We randomly choose 10 images from COCO minival and calculate the L2 distance of features in different components of different training strategies. As shown in Figure \ref{figure4}, “All” denotes that the detector with full data training; ‘Incremental’ denotes that the detector with incremental data training; “Finetune” denotes that the detector with finetuning training. Distilling student detector via classification-based and regression-based incremental response in the head can substantially narrow the distances with upper bound. However, neither the L2 distance between “All vs. Incremental” and “All vs. Finetune” improves significantly in the FPN representing the feature-based distillation. This also supports our assumption that different response from the head has its particularly significant, especially classification response. 

\textbf{Incremental response helps both learning and generalization.} We notice that the incremental response from the head can provide an effective guidance to avoid catastrophic forgetting problems. Thereby, the student detector achieves noticeable improvement in different scenarios. In Table \ref{table4}, our method can still learn new object classes without forgetting old ones even with a little data. But, due to the insufficient incremental response provided in the +5 classes scenario, our method did not achieve a more competitive AP. However, our method still contributes to generalization. In this case, we can degrade the adaptive response to all responses in exchange for a better compromise. Comparatively, when sufficient incremental responses emerge, our method is easy to achieve (near) perfect AP.

\begin{table}[]
\centering
\caption{Ablation study based on GFLV1 detector using the COCO benchmark under first 40 classes + last 40 classes. ("$\Delta$" represents an improvement over Catastrophic Forgetting. "$\nabla$" represents the gap with Upper Bound.)}
\label{table3}
\begin{tabular}{@{}l|cccccc@{}}
\toprule
Method & $AP$ & $AP_{50}$ & $AP_{75}$  & $AP_{S}$  & $AP_{M}$  & $AP_{L}$  \\ \midrule
Upper Bound & 40.2 & 58.3  & 43.6  & 23.2  & 44.1   & 52.2  \\
Catastrophic Forgetting & 17.8 & 25.9  & 19.3 & 8.3  & 19.2 & 24.6\\ \midrule
KD:all cls + all reg & 31.5($\Delta 13.7/\nabla 8.7$) & 48.3  & 33.4 & 17.7  & 35.3 & 41.3\\
KD:all cls & 23.8($\Delta 6.0/\nabla 16.4$) & 36.6  & 24.9 & 11.8  & 27.2 & 32.9\\
KD:all reg & 13.0($\Delta -4.8 /\nabla 27.2$) & 21.1  & 13.4 & 5.0  & 14.7 & 18.6\\
KD:cls + APS & 33.2($\Delta 15.4/\nabla 7.0$) & 51.2  & 35.2 & 18.5  & 37.8 & 43.8\\
KD:cls + reg + APS & \textbf{36.9($\Delta 19.1/\nabla 3.3$)} & \textbf{54.5}  & \textbf{39.6}  & \textbf{21.3}  & \textbf{40.4}   & \textbf{47.5}\\ \bottomrule
\end{tabular}
\end{table}

\begin{table}[]
\centering
\caption{Incremental results based on GFLV1 detector on COCO benchmark under first 75 classes + last 5 classes. ("$\Delta$" represents an improvement over Catastrophic Forgetting.)}
\label{table4}
\begin{tabular}{@{}l|cccccc@{}}
\toprule
Method & $AP$ & $AP_{50}$ & $AP_{75}$  & $AP_{S}$  & $AP_{M}$  & $AP_{L}$  \\ \midrule
Catastrophic Forgetting & 3.8 & 5.9  & 3.8 & 1.9  & 5.3 & 6.5\\ \midrule
All response & 32.5 ($\Delta 28.7$) & 48.9  & 34.7  & 18.3  & 35.9   & 41.1  \\
Adaptive response  & 28.3 ($\Delta 24.5$) & 42.4  & 30.3  & 15.0  & 31.5   & 37.0\\ \bottomrule
\end{tabular}
\end{table}

\section{Conclusion}

In this paper, we design an entirely response-based incremental object detection paradigm. This method uses only the detection head to achieve incremental detection, which significantly alleviates catastrophic forgetting. We innovatively learn responses from detection bounding boxes and classification predictions, and specifically introduce incremental localization distillation in the regression response. Second, the adaptive selection technique is designed to provide a fair incremental response in the different heads. Extensive experiments validate the effectiveness of our method. Finally, our empirical analysis reveals the contribution of different responses and components in incremental detection, which could provide insights to further advancement in the field.

\bibliography{iclr2022_conference}
\bibliographystyle{iclr2022_conference}

\end{document}